\documentclass[conference]{IEEEtran}
\IEEEoverridecommandlockouts
\usepackage{cite}
\usepackage{amsmath,amssymb,amsfonts}
\usepackage{algorithm}
\usepackage{pifont}
\usepackage{algorithmic}
\usepackage{graphicx}
\usepackage{float}
\usepackage{textcomp}
\usepackage{url}
\usepackage[table]{xcolor}
\def\BibTeX{{\rm B\kern-.05em{\sc i\kern-.025em b}\kern-.08em
    T\kern-.1667em\lower.7ex\hbox{E}\kern-.125emX}}
\newcommand{\xgbcell}[1]{\cellcolor{black!8}\textcolor{black!65}{#1}}
\begin{document}

\title{Adaptive Multi-Branching for \\ Shallow Decision Tree Induction
}

\author{
\IEEEauthorblockN{
Hanul Park\textsuperscript{1},
Jeonghoon Choi\textsuperscript{1},
Juseong Kim\textsuperscript{1},
Sanghun Sel\textsuperscript{1},
Giltae Song\textsuperscript{1,2,*}
}
\IEEEauthorblockA{
\textsuperscript{1}Department of Information Convergence Engineering, Pusan National University, Busan, Rep. of Korea\\
\textsuperscript{2}School of Computer Science and Engineering, Pusan National University, Busan, Rep. of Korea\\
\textsuperscript{*}Corresponding author: \texttt{gsong@pusan.ac.kr}\\
\texttt{hanul.park@gmail.com, jeonghoonchoi@pusan.ac.kr}\\
\texttt{kjs\_0322@naver.com, poohnim5@naver.com}
}
}

\maketitle

\begingroup
\renewcommand{\thefootnote}{}
\footnotetext{\footnotesize
\textcopyright~2026 IEEE. Personal use of this material is permitted.
Permission from IEEE must be obtained for all other uses, in any current or
future media, including reprinting/republishing this material for advertising
or promotional purposes, creating new collective works, for resale or
redistribution to servers or lists, or reuse of any copyrighted component of
this work in other works.}
\endgroup

\begin{abstract}
Decision trees are attractive for tabular prediction tasks because each prediction follows an interpretable sequence of feature-threshold tests. Under a strict maximum-depth budget, however, conventional binary trees can be under-expressive, since each internal node makes only a single threshold decision. We study shallow-depth tree induction, where the goal is to improve accuracy while keeping root-to-leaf paths short. We propose the Multi-Branch Neural Decision Tree with Adaptive Pruning (MBNDT), a single axis-aligned tree trained end-to-end with differentiable multi-way splits. Each internal node learns ordered thresholds over a selected feature and a branch mask that adapts its effective arity, and the trained model is converted to a deterministic single-path tree for inference. Across 21 OpenML binary-classification benchmarks, MBNDT achieves the best average rank and mean balanced accuracy among depth-constrained single-tree baselines; a controlled ablation isolates multi-way splitting as the source of the gain. These gains come with an explicit trade-off: MBNDT realizes more leaves than the other single-tree baselines, making it best suited when accuracy under short, bounded decision paths is prioritized over minimal global tree size. 
\end{abstract}


\section{Introduction}
\label{submission}

Decision trees remain a central model class for tabular prediction tasks because of their intrinsic interpretability. Unlike black-box models, a decision tree maps each input to a deterministic root-to-leaf path consisting of feature-threshold tests, allowing individual predictions to be inspected as rule-like decision processes. Accordingly, this path-based structure is particularly attractive in settings where predictions must be explained, audited, or implemented through a small number of sequential decisions \cite{loh2014fifty, huysmans2011empirical, rudin2019stop}.

Although decision trees are commonly regarded as interpretable models, their interpretability depends on human cognitive factors and the intended use context; no single global complexity measure can fully capture it \cite{freitas2014comprehensible, piltaver2016what, huysmans2011empirical}. For example, a tree may contain many leaves while assigning each instance through a short path; conversely, a globally compact tree may contain paths that are too deep for practical inspection. Because predictions are typically interpreted one instance at a time, path length is considered as a proxy for inspection effort while reporting leaf count to expose the corresponding global-size trade-off \cite{piltaver2016what, souza2022decision}. Shallow trees are therefore especially relevant in settings where predictions must be manually reviewed, audited, or implemented as rule-like decision protocols, such as high-stakes decision support and clinical risk stratification \cite{rudin2019stop, letham2015interpretable}. In this paper, we study \emph{shallow-depth tree induction}, where the maximum decision depth is strictly constrained, e.g., $D \leq 4$. In this regime, the goal is not merely to reduce the total number of leaves, but to improve predictive accuracy under a short per-instance decision-path budget.

Conventional binary decision trees, however, face an expressivity bottleneck in this shallow-depth regime. A binary tree of depth $D$ can represent at most $2^D$ leaves, and each internal node partitions its region using only a single threshold decision. When $D$ is small, this coupling between path length and partitioning capacity can make a single binary tree under-expressive for heterogeneous tabular decision boundaries. Specifically, greedy methods \cite{breiman1984cart, quinlan1993c45} make locally suboptimal root-level decisions that are difficult to correct under a small depth budget; solver-based methods improve global search but typically operate over binary or restricted candidate split structures \cite{bertsimas2017oct}; and differentiable tree methods often retain binary routing, soft prediction, or ensemble-style formulations that are hard to interpret.

Our key idea is to improve the local expressivity of each decision step by dividing a selected feature into multiple ordered intervals using a multi-way split. We propose the \emph{Multi-Branch Neural Decision Tree with Adaptive Pruning} (MBNDT)\footnote{Implementation: [\url{https://github.com/hanulpark98/MBNDT}]}, an axis-aligned shallow decision tree trained end-to-end by gradient-based optimization. Each internal node learns a feature selector and a set of strictly ordered thresholds that define a differentiable multi-way split over the selected feature. Moreover, MBNDT avoids node-wise greedy split selection by jointly optimizing split features, thresholds and leaf predictions. To avoid unnecessarily using the full branching factor at every node, MBNDT introduces learnable branch masks that adapt the effective arity of each split during training. After training, the learned routing structure and branch masks are converted into a deterministic pruned tree for inference, yielding a single root-to-leaf prediction path for each instance.

Our empirical study evaluates MBNDT against greedy, solver-based, and
gradient-based single-tree baselines under fixed shallow-depth budgets. Across
the evaluated datasets, MBNDT achieves the best average rank and highest mean
balanced accuracy among the depth-constrained single-tree learners, while
sparsification ablations show that branch masks and post-hoc pruning
reduce realized leaves without degrading balanced accuracy. These
gains carry an explicit complexity trade-off---MBNDT realizes more leaves than
the other single-tree baselines---so it is best suited to settings that
prioritize accuracy under short per-instance decision paths over minimal global
tree size.

Our contributions are as follows:
\begin{itemize}
\item \textbf{Adaptive multi-way branching.} We introduce differentiable,
axis-aligned multi-way interval splits: each internal node selects a single
feature and learns ordered thresholds that partition it into several regions,
with split features, thresholds, and leaf predictions optimized jointly by
gradient descent rather than greedily.

\item \textbf{Structure sparsification for differentiable trees.} We adapt
each node's effective arity with learnable branch masks, control tree size
during training through a differentiable leaf-budget penalty, and apply
post-hoc train-path pruning---together converting the nominal $B$-ary tree
into a compact effective tree with one deterministic path per input.

\item \textbf{Accuracy--depth--size analysis.} We benchmark MBNDT against
greedy, solver-based, and gradient-based single-tree baselines under a shared
depth budget, reporting predictive accuracy, decision-path length, and
realized leaves, with ablations isolating multi-way branching and the
sparsification mechanisms.
\end{itemize}

\section{Related Work}

\subsection{Rule Simplicity and Explanation Redundancy}
Several studies analyze tree interpretability through the complexity of individual root-to-leaf rules. Souza et al. \cite{souza2022decision} define explanation size as the number of distinct attributes appearing along a decision path. A complete path, however, need not be a minimal explanation: Izza et al. \cite{izza2022redundancy} study redundant path conditions, while McTavish et al. \cite{mctavish2025predictive} show that predictively equivalent trees can induce different evaluation processes. These concerns are complementary to our scope: we bound the number of executed decision steps rather than claim subset-minimal explanations.

\subsection{Greedy Decision Tree Induction}

Classical decision tree learners typically use a top-down recursive partitioning strategy: at each node, they select the feature and threshold that maximizes a local impurity reduction, such as information gain or the Gini index, and then recurse on the resulting children. Representative algorithms include ID3 \cite{quinlan1986induction}, C4.5 \cite{quinlan1993c45}, and CART \cite{breiman1984cart}. These methods are fast and scalable, but optimizing each split in isolation can produce globally suboptimal trees, a limitation that is particularly consequential under shallow depth budgets \cite{bertsimas2017oct}.

\subsection{Optimal Decision Tree Search}

Beyond greedy induction, optimal-tree methods formulate tree learning as a discrete global optimization problem, typically under fixed depth, leaf, or feature budgets. Mixed-integer programming approaches encode routing, split selection, and leaf predictions and solve the resulting formulation using branch-and-bound \cite{bertsimas2017oct}. Specialized methods improve scalability through dynamic programming, bounds, and search reuse \cite{hu2019optimalsparse,aglin2021pydl85,demirovic2022murtree,lin2020generalized,vanderlinden2023streed}. Recent work also reduces reliance on coarse discretization: ConTree directly optimizes continuous-feature thresholds using dynamic programming with branch-and-bound \cite{brita2025contree}, while SPLIT combines bounded lookahead with greedy lower-level splits for near-optimal search \cite{babbar2025split}.

\subsection{Multi-way Splitting for Decision Trees}
Multi-way splits have long been used for categorical attributes in ID3/C4.5-style and CHAID trees \cite{quinlan1986induction,quinlan1993c45,kass1980chaid}. Numerical variants search for multiple thresholds at each node to obtain smaller or more expressive trees \cite{fulton1995multiway,berzal2004multiway}. More recently, path-based mixed-integer formulations with column generation have been used to learn constrained optimal multiway-split trees \cite{subramanian2023multiway}. The earlier node-wise methods, however, do not optimize the structure jointly across levels, and higher arity can fragment the data and increase overfitting, potentially requiring post-hoc pruning.

\subsection{Differentiable Decision Trees}

Another line of work replaces node-wise greedy induction with gradient-based optimization. These methods replace discrete split selection with differentiable relaxations or surrogate gradients, enabling tree parameters to be optimized by backpropagation. Good et al.\cite{good2023feature} alternate sparse feature learning with differentiable tree construction to obtain compact trees. Norouzi et al.\cite{norouzi2015nongreedy} jointly optimize oblique splits across levels and leaf parameters, while DTSemNet\cite{panda2024vanilla} learns hard oblique trees through a neural-network encoding. Soft-routing methods use differentiable path probabilities \cite{kontschieder2015dndf,hehn2020endtoend}. Particularly relevant, DNDT learns feature-wise multi-interval cut points by gradient descent, and D3T adapts the number of cut points per feature \cite{yang2018deepneuraldt,zhou2025hademif}. These methods discretize features globally, whereas MBNDT places adaptive multi-way splits at internal nodes of a recursive tree. GradTree instead learns hard axis-aligned trees through surrogate gradients, while GRANDE extends gradient-trained trees to ensembles \cite{marton2024gradtree,marton2024grande}. MBNDT further converts the learned structure into a deterministic pruned tree for single-path inference.

\begin{figure*}[!t] % or [tb]; with dblfloatfix you can try [b]
  \centering
  \includegraphics[width=\textwidth]{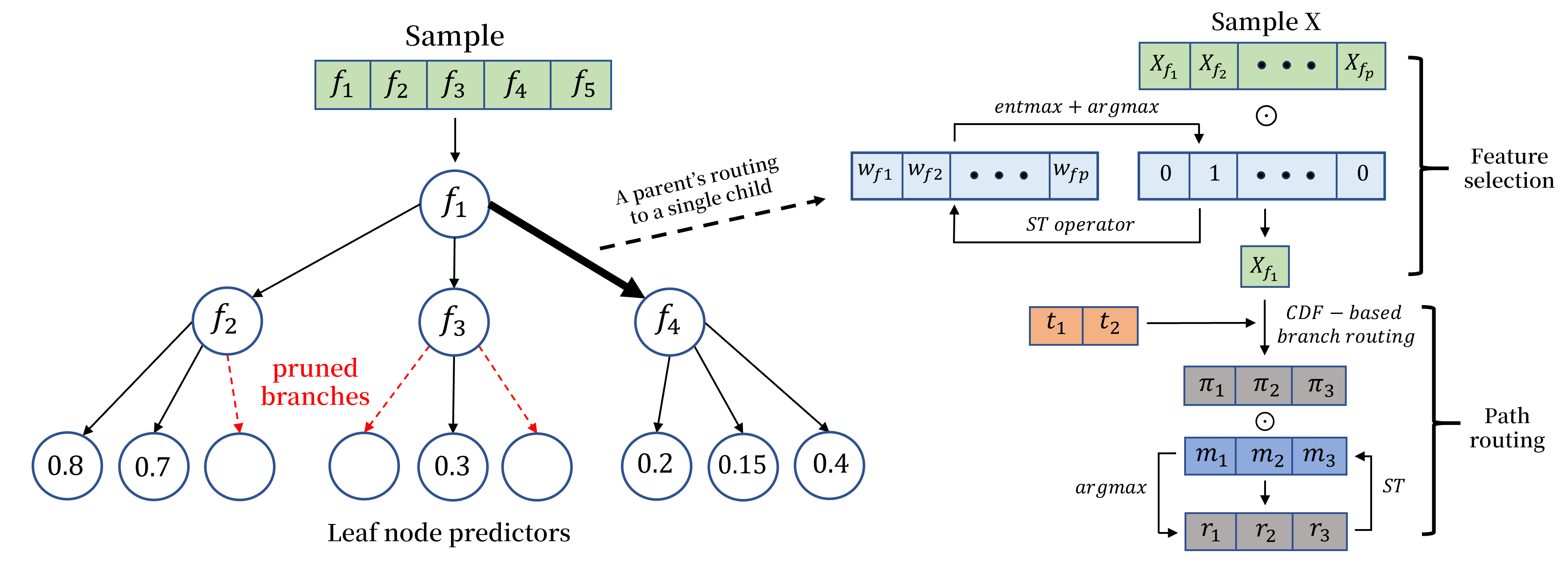}

  \caption{Overview of MBNDT with prespecified maximum branching factor $B=3$. \textbf{Left}: a nominal depth-2 ternary tree with mask-suppressed branches. \textbf{Right}: node-level routing, where a selected feature is partitioned by ordered thresholds into soft interval probabilities, reweighted by branch masks, and hard-routed to a single child.}

  \label{fig:MBNDT-routing}
\end{figure*}

\subsection{Positioning}
The individual ingredients of MBNDT are not new in isolation: multi-way
numerical splits, gradient-based tree training, and global threshold
optimization have each been studied separately above. Our contribution is
their unification into a decision tree learner that learns ordered
\emph{multi-way} interval splits with continuous thresholds and adapts each
node's \emph{effective arity}, while remaining a single axis-aligned tree.

\section{MBNDT: Architecture, Training, and Inference}
\label{sec:method}

We formulate the learning of \textbf{MBNDT} as the joint optimization of the parameters $\theta$ of a predefined decision tree $T_{\theta}$ with branching factor $B$ at each internal node and depth $D$. An axis-aligned node selects one input feature, and routing denotes assigning an input to one child according to the interval containing that feature value. The learnable parameters consist of node-wise split feature parameters, split threshold parameters, branch masks, and $B^D$ leaf logits. Figure~\ref{fig:MBNDT-routing} illustrates the node-level pipeline.

Given a labeled training set $\mathcal{S}=\{(\mathbf{x}_i,y_i)\}_{i=1}^{n}$ with $\mathbf{x}_i\in\mathbb{R}^{p}$ and binary label $y_i\in\{0,1\}$, MBNDT outputs a scalar logit $f_{\theta}(\mathbf{x})\in\mathbb{R}$. Its sigmoid transform $\sigma(f_{\theta}(\mathbf{x}))$ is interpreted as the estimated probability of the positive class. In our experiments, we train the model using binary cross-entropy together with a leaf-budget regularization term defined as:
\begin{equation}
\label{eq:train_obj}
\min_{\theta}\;
\frac{1}{n}\sum_{i=1}^{n}
\mathcal{L}_{\mathrm{BCE}}\!\left(y_i, f_{\theta}(\mathbf{x}_i)\right)
\;+\;
\mathcal{L}_{\mathrm{budget}}(\theta),
\end{equation}
We defer the detailed form of $\mathcal{L}_{\mathrm{budget}}$ to Section~\ref{sec:4.2}.

\subsection{Split Feature Selection}
\label{sec:3.1}

At internal node $j$, we associate node $j$ with a vector of learnable \emph{feature logits} $\mathbf{s}_{j}\in\mathbb{R}^p$, whose $k$-th entry $s_{j,k}$ represents the tendency to select feature $k$ at node $j$. These logits are then converted into a sparse probability vector over features using $\alpha$-entmax \cite{peters2019sparses2s}:
\begin{equation}
\mathbf{w}_j \;=\; \operatorname{entmax}_{1.5}\!\left(\mathbf{s}_j\right).
\label{eq:feature_entmax}
\end{equation}
Unlike other relaxation functions (e.g., softmax, Gumbel--Softmax), entmax can assign \emph{exact zeros} to low-scoring features, yielding a highly sparse candidate set per node while remaining differentiable; this sparsity is often empirically associated with more stable optimization and improved interpretability in differentiable routing/tree models \cite{marton2024gradtree, popov2019node}.

To guarantee \emph{axis-aligned} splits, node $j$ must ultimately select a \emph{single} feature. 
We therefore harden $\mathbf{w}_j$ via $\arg\max$ in the forward pass (as a one-hot indicator) and use a straight-through (ST) estimator
\cite{bengio2013stochasticneurons} for backpropagation, following the prior studies \cite{marton2024gradtree,karthikeyanlearning}:
\begin{equation}
\tilde{\mathbf{w}}_j
\;=\;
\mathbf{w}_j \;+\; \mathrm{sg}\!\Big(\mathrm{onehot}(\arg\max_k\, w_{j,k})-\mathbf{w}_j\Big),
\label{eq:st_onehot}
\end{equation}
where $\mathrm{sg}(\cdot)$ denotes the stop-gradient operator.
This construction uses the hard one-hot choice in the forward pass, while in the backward pass it propagates gradients as if
$\tilde{\mathbf{w}}_j=\mathbf{w}_j$, computed via the $\mathrm{entmax}_{1.5}$ Jacobian:
\begin{equation}
\frac{\partial \tilde{\mathbf{w}}_j}{\partial \mathbf{s}_j}
\;\approx\;
\frac{\partial \mathbf{w}_j}{\partial \mathbf{s}_j}
\;=\;
J_{\mathrm{entmax}_{1.5}}(\mathbf{s}_j).
\label{eq:entmax_jacobian}
\end{equation}

\subsection{Multi-branch routing via ordered thresholds}
\label{sec:3.2}

A standard binary decision-tree node learns a single threshold and partitions its input into two regions. In contrast, each internal node of MBNDT learns $B{-}1$ \emph{ordered} thresholds and partitions the corresponding node score into $B$ ordered regions. This requires the thresholds $t_{j,1}<\cdots<t_{j,B-1}$ to be strictly increasing. Rather than enforcing this constraint via per-iteration sorting, we parameterize the thresholds as cumulative sums of $\operatorname{softplus}$-transformed learnable gap logits $\delta_{j,r}$. For node $j$, given a learnable base threshold $t_{j,0}$, the ordered thresholds are defined as

\begin{equation}
g_{j,r}=\operatorname{softplus}(\delta_{j,r})>0,\quad \boldsymbol{\delta}_j\in\mathbb{R}^{B-1},
\end{equation}
\begin{equation}
t_{j,k}=t_{j,0}+\sum_{r=1}^{k} g_{j,r},\quad k=1,\dots,B-1,
\end{equation}

which ensures $t_{j,1}<\cdots<t_{j,B-1}$ by construction. The $\operatorname{softplus}(x)=\log(1+e^x)$ transformation maps unconstrained real-valued parameters to strictly positive gaps, thereby preventing coincident thresholds while remaining smooth for gradient-based optimization. 

Given the ordered thresholds $\{t_{j,k}\}_{k=1}^{B-1}$, node $j$ routes an input $\mathbf{x}$ according to its selected-feature score $z_j(\mathbf{x})=\mathbf{x}^{\top}\tilde{\mathbf{w}}_j$. Because hard assignment of $z_j(\mathbf{x})$ to one of the $B$ intervals is non-differentiable, we instead use a soft binning scheme based on cumulative logistic probabilities. For a temperature $\tau_{\mathrm{cdf}}>0$, we define

\begin{equation}
u_{j,k}(\mathbf{x}) \;=\; \frac{t_{j,k}-z_j(\mathbf{x})}{\tau_{\mathrm{cdf}}}, \quad k=1,\dots,B-1,
\label{eq:uk}
\end{equation}
and define the cumulative terms as $C_{j,k}(\mathbf{x})=\sigma(u_{j,k}(\mathbf{x}))$, where $\sigma(u)=1/(1+e^{-u})$.
The $B$ soft routing probabilities are then obtained from adjacent differences:
\begin{equation}
\label{eq:logit_bins}
\begin{split}
\pi_{j,1}(\mathbf{x}) &= C_{j,1}(\mathbf{x}),\\
\pi_{j,k}(\mathbf{x}) &= C_{j,k}(\mathbf{x})-C_{j,k-1}(\mathbf{x}),
\qquad k=2,\dots,B-1,\\
\pi_{j,B}(\mathbf{x}) &= 1-C_{j,B-1}(\mathbf{x}).
\end{split}
\end{equation}

This defines a differentiable approximation to interval membership that becomes sharper as $\tau_{\mathrm{cdf}}\downarrow 0$. Since $\sigma$ is monotone increasing and $t_{j,1}<\cdots<t_{j,B-1}$, we have
$C_{j,1}(\mathbf{x})\le\cdots\le C_{j,B-1}(\mathbf{x})$, implying $\pi_{j,k}(\mathbf{x})\ge 0$ and $\sum_{k=1}^{B}\pi_{j,k}(\mathbf{x})=1$. Thus, $\boldsymbol{\pi}_j(\mathbf{x})$ defines a valid routing distribution over the $B$ branches.

\subsection{Adaptive pruning via branch masks}
\label{sec:3.3}

A full $B$-ary tree of depth $D$ grows exponentially in size, even though many internal nodes may not require all $B$ outgoing branches to form an effective partition. To allow each node to adapt its effective arity, MBNDT equips every internal node $j$ with a learnable, input-independent branch mask over its $B$ outgoing branches. Thus, $B$ serves only as an upper bound on the local branching factor, and unnecessary branches can be suppressed during training.

Specifically, for node $j$, we introduce sigmoid-transformed branch masks $\mathbf{m}_j=\sigma(\boldsymbol{\eta}_j)$, $\boldsymbol{\eta}_j\in\mathbb{R}^{B}$. Given the unmasked routing probabilities $\pi_{j,b}(\mathbf{x})$ from Equation~\eqref{eq:logit_bins}, the normalized masked routing probabilities become
\begin{equation}
\tilde{\pi}_{j,b}(\mathbf{x})
=
\frac{\mathbf{m}_{j,b}\,\pi_{j,b}(\mathbf{x})}
{\sum_{c=1}^{B} \mathbf{m}_{j,c}\,\pi_{j,c}(\mathbf{x})},
\quad b=1,\dots,B,
\label{eq:branch-mask}
\end{equation}
This preserves the partition-of-unity constraint while allowing node $j$ to globally suppress or emphasize individual branches. Using the masked probabilities $\tilde{\boldsymbol{\pi}}_j(\mathbf{x})$, we perform hard routing in the forward pass by selecting the branch with highest $\tilde{\boldsymbol{\pi}}_j(\mathbf{x})$ and use a straight-through estimator:
\begin{equation}
\mathbf{r}_j(\mathbf{x})
=
\tilde{\boldsymbol{\pi}}_j(\mathbf{x})
+\mathrm{sg}\!\Big(
\mathrm{onehot}\!\big(\arg\max_k \tilde{\pi}_{j,k}(\mathbf{x})\big)
-\tilde{\boldsymbol{\pi}}_j(\mathbf{x})
\Big),
\label{eq:branch-mask-st}
\end{equation}
so that $\mathbf{r}_j(\mathbf{x})$ is one-hot in the forward pass, while gradients are propagated through $\tilde{\boldsymbol{\pi}}_j(\mathbf{x})$ during backpropagation.

\begin{figure}[t]
  \centering
  \includegraphics[width=\columnwidth]{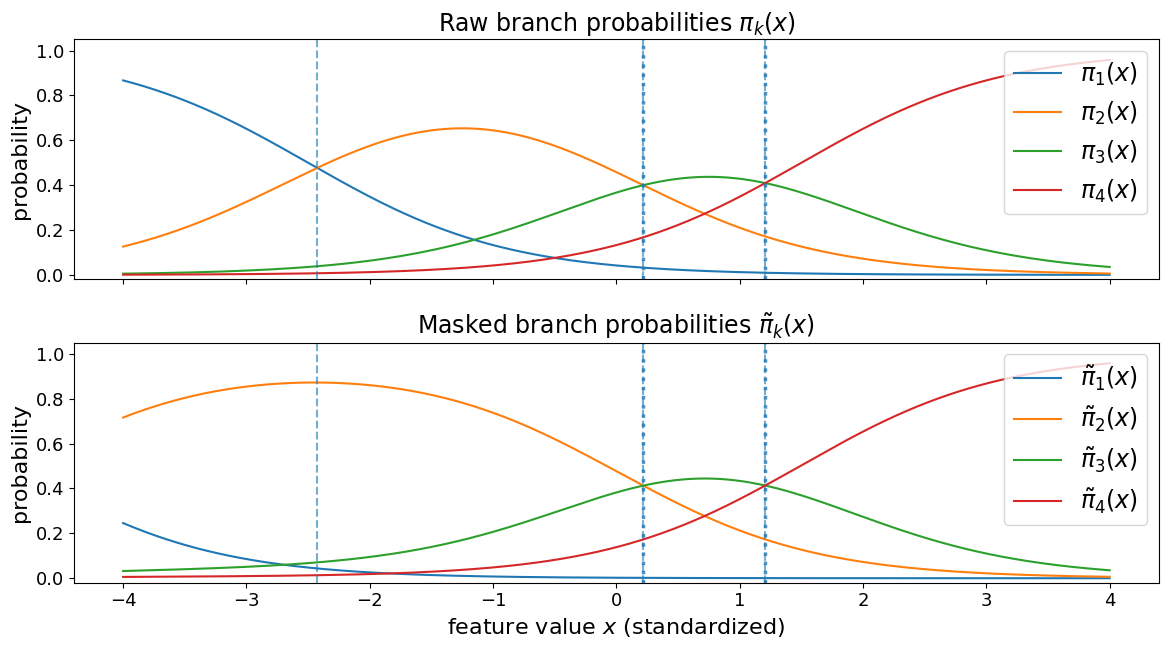}
  \caption{Illustration of the branch-mask effect in a toy $B{=}4$ routing node. \textbf{Top}: raw interval probabilities; \textbf{Bottom}: probabilities after applying branch masks and renormalizing. Dashed lines show the original ordered-threshold locations; masks reweight branch probabilities and can change which branches are selected under hard routing.}
  \label{fig:mask_effect}
\end{figure}

\paragraph{Pruning effect.}
The branch masks directly induce adaptive pruning under hard routing. If
$m_{j,k}\approx 0$, then $\tilde{\pi}_{j,k}(\mathbf{x})\approx 0$ for all
$\mathbf{x}$, so branch $k$ is effectively never selected and its subtree
becomes unreachable at inference. More generally, the masks modify pairwise
branch preferences through multiplicative biases:
\begin{equation}
\frac{\tilde{\pi}_{j,k}(\mathbf{x})}{\tilde{\pi}_{j,\ell}(\mathbf{x})}
=
\frac{\mathbf{m}_{j,k}}{\mathbf{m}_{j,\ell}}
\cdot
\frac{\pi_{j,k}(\mathbf{x})}{\pi_{j,\ell}(\mathbf{x})},
\label{eq:pruning-effect}
\end{equation}
which implies $\log\frac{\tilde{\pi}_{j,k}(\mathbf{x})}{\tilde{\pi}_{j,\ell}(\mathbf{x})}
=
\log\frac{\pi_{j,k}(\mathbf{x})}{\pi_{j,\ell}(\mathbf{x})}
+
\log\frac{\mathbf{m}_{j,k}}{\mathbf{m}_{j,\ell}}$. Thus, the masks shift hard-routing preferences by branch-specific global offsets, allowing weak branches to be suppressed across the input space.
Figure~\ref{fig:mask_effect} contrasts unmasked routing probabilities $\pi_k(x)$ with masked probabilities $\tilde{\pi}_k(x)$, showing how masks reweight branch preferences and suppress weak branches while the ordered thresholds remain fixed.

\subsection{Inference}
\label{sec:3.4}

Each leaf node $\ell$ stores a scalar logit $\theta_\ell \in \mathbb{R}$. Since only leaf nodes store learned output logits, MBNDT computes predictions by a bottom-up reduction from the leaves to the root. Specifically, for each leaf $\ell$, we initialize the subtree value as $v^{(D)}_{\ell}(\mathbf{x})=\theta_\ell$. Internal nodes do not store output logits; instead, they are assigned recursively computed subtree values. Let $\mathrm{child}(j,k)$ denote the $k$-th child of node $j$. For depths $d=D-1,\dots,0$, we compute
\begin{equation}
v^{(d)}_{j}(\mathbf{x})
=
\sum_{k=1}^{B} r_{j,k}(\mathbf{x})\, v^{(d+1)}_{\mathrm{child}(j,k)}(\mathbf{x}).
\label{eq:bottomup}
\end{equation}
The tree logit is then defined as $f_{\theta}(\mathbf{x})=v^{(0)}_{\mathrm{root}}(\mathbf{x})$. At inference, $\mathbf r_j(\mathbf x)\in\{0,1\}^B$ is one-hot, so exactly one term in the sum is active at each internal node. Therefore, the recursion simply propagates the logit of the uniquely selected leaf upward to the root. For binary classification, the predicted probability is $\sigma(f_{\theta}(\mathbf{x}))$.

\section{Structure Sparsification in MBNDT}
\label{sec:4}

MBNDT supports structure sparsification at three levels: intrinsic sparsification through learned branch masks, explicit complexity control through leaf-budget regularization, and additional compression through post-hoc train-path pruning.

\subsection{Effective sparsification from branch masks}
\label{sec:4.1}

As introduced in Section~\ref{sec:3.3}, the branch masks multiplicatively bias routing probabilities and can suppress weak branches under hard routing. This reduces the effective arity of internal nodes and may render entire subtrees unreachable. We therefore distinguish the \emph{nominal tree} from the \emph{effective tree}, obtained by retaining only branches that remain reachable under hard routing. A leaf is called \emph{routable} if there exists an admissible input $\mathbf{x}$ whose hard routing decisions reach that leaf. The effective tree is the subtree induced by all routable leaves.

\subsection{Leaf-budget regularization}
\label{sec:4.2}

To further control tree complexity during training, we introduce an augmented-Lagrangian style \cite{burman2023augmented} penalty for exceeding a predefined leaf budget. Since the exact number of hard-routable leaves is discrete and non-differentiable, we instead construct a differentiable surrogate from the masked soft routing probabilities in Equation~\eqref{eq:branch-mask}.

Let $\tilde{\pi}_{j,b}(\mathbf{x})$ denote the masked routing probability of branch $b$ at internal node $j$, and let $\bar{\tilde{\pi}}_{j,b}
=
\mathbb{E}_{\mathbf{x}}\!\left[\tilde{\pi}_{j,b}(\mathbf{x})\right]$
denote its average over the current batch. Motivated by effective-number measures such as the inverse Simpson index \cite{chao2014rarefaction}, we define the effective arity of node $j$ as
\begin{equation}
\mathrm{Neff}_j
=
\frac{1}{\sum_{b=1}^{B} \bar{\tilde{\pi}}_{j,b}^{\,2}},
\label{eq:neff}
\end{equation}
which is close to $1$ when routing concentrates on a single branch and increases toward $B$ as branch usage becomes more distributed. We then calculate the soft reach probability $s_j(\mathbf{x})$ of node $j$, defined recursively by $s_{\mathrm{root}}(\mathbf{x})=1$ and $s_{\mathrm{child}(j,b)}(\mathbf{x})=s_j(\mathbf{x})\tilde{\pi}_{j,b}(\mathbf{x})$. Consequently, we use $\alpha_j=\mathbb{E}_{\mathbf{x}}[s_j(\mathbf{x})]$ and define the soft effective leaf count as
\begin{equation}
L_{\mathrm{soft}}
=
1+\sum_j \alpha_j \bigl(\mathrm{Neff}_j-1\bigr).
\label{eq:lsoft}
\end{equation}
Although $L_{\mathrm{soft}}$ is not identical to the exact hard-routable leaf count, it provides a differentiable proxy for effective structural usage during optimization. Thus, given a target leaf budget $K$ we penalize budget violations using
\begin{equation}
\mathcal{L}_{\mathrm{budget}}
=
\mu [L_{\mathrm{soft}}-K]_+
+
\frac{\rho}{2}[L_{\mathrm{soft}}-K]_+^2,
\label{eq:leafbudget}
\end{equation}
where $[z]_+=\max(z,0)$ so sparse trees are not penalized. The update schedule for the coefficients $\rho$ and $\mu$ is described in the Appendix.

\subsection{Post-hoc train-path pruning}
\label{sec:4.3}

In addition to training-time sparsification, MBNDT applies post-hoc train-path pruning (PP) based on the hard-routing behavior of the trained model on the training set. At each internal node, we retain only branches that are observed at least once under hard routing, thereby obtaining a train-path-supported subtree. Inputs that would otherwise select a removed branch are redirected to the nearest surviving branch in the original branch ordering, while the original leaf logits are kept unchanged. This pruned tree is the final MBNDT model used at inference, rather than a newly retrained compact tree.

\begin{algorithm}[t]
\caption{Training and construction of MBNDT}
\label{alg:mbndt}
\small
\begin{algorithmic}[1]
\REQUIRE Training set $\mathcal{S}$, depth $D$, maximum arity $B$, leaf budget $K$
\STATE Initialize a full $B$-ary tree $T_\theta$
\FOR{each training minibatch $\mathcal{B}$}
    \FOR{each internal node $j$}
        \STATE Select one feature using the ST feature selector
        \STATE Construct $B{-}1$ ordered thresholds
        \STATE Compute masked branch probabilities and ST hard routing
    \ENDFOR
    \STATE Route each sample to a leaf and obtain $f_\theta(\mathbf{x})$
    \STATE Update $\theta$ using
    $\mathcal{L}_{\mathrm{BCE}}+\mathcal{L}_{\mathrm{budget}}$
\ENDFOR
\STATE Determine branches used by the training samples
\STATE Prune unused branches and redirect them to the nearest surviving branch
\STATE Fix feature and branch choices by their hard decisions
\RETURN Deterministic pruned tree $T^\star$
\end{algorithmic}
\end{algorithm}

\section{Experiments}
\label{sec:5}

\subsection{Experimental setting}

\paragraph{Baselines}

Our single-tree baselines are CART~\cite{breiman1984cart} (greedy binary tree),
SPLIT~\cite{babbar2025split} (solver-based shallow tree), and GradTree~\cite{marton2024gradtree}
(gradient-trained tree). We restrict the comparison to methods that, like
MBNDT, produce a single axis-aligned tree with one root-to-leaf path per
input; this excludes soft and ensemble differentiable trees, whose predictions are not single-path. We additionally report XGBoost~\cite{chen2016xgboost} as a strong black-box reference, but do not
treat it as a comparable single-tree model, since its predictions aggregate
many trees rather than following one decision path. 

Exact and near-optimal tree learners such as OSDT, MurTree, and
GOSDT~\cite{hu2019optimalsparse, demirovic2022murtree, lin2020generalized} are comparable in form. However, they operate
over pre-binarized feature sets, and their search cost grows steeply with the
number of binarized features and training instances, making per-split
hyperparameter tuning under our shared one-hour budget impractical on the
larger benchmarks (up to ${\sim}10^6$ instances). We therefore adopt
SPLIT~\cite{babbar2025split} as the representative of this family: its
original study reports near-optimal tree quality while substantially reducing
search time relative to exact optimal-tree solvers, making it the closest
scalable proxy for exact and near-optimal induction across all dataset sizes
rather than only on the small datasets where exact solvers remain tractable. 

\begin{table*}[t]
\caption{Per-dataset test balanced accuracy on 21 binary-classification datasets.
Values are reported as mean $\pm$ standard deviation over five dataset splits.
All single-tree methods use a maximum-depth budget of $D\leq4$; XGBoost is
included as a depth-limited ensemble reference. Ties are assigned the same competition rank and counted as wins when tied for first.}
\label{tab:per_dataset_balacc}
\centering
\scriptsize
\setlength{\tabcolsep}{6pt}
\renewcommand{\arraystretch}{0.92}
\resizebox{\textwidth}{!}{%
\begin{tabular}{l c c c c c}
\hline
Dataset & \multicolumn{1}{c}{\xgbcell{Ensemble (ref.)}} & \multicolumn{4}{c}{Single tree} \\
\cline{2-2}\cline{3-6}
 & \xgbcell{XGBoost} & \multicolumn{2}{c}{Gradient-based} & Solver-based & Conventional \\
\cline{3-4}\cline{5-5}\cline{6-6}
 & \xgbcell{} & MBNDT (ours) & GradTree & SPLIT & CART \\
\hline
\multicolumn{6}{l}{\textbf{Small datasets ($n < 10^3$)}} \\
hepatitis & \xgbcell{0.745 $\pm$ 0.113} & \textbf{0.675 $\pm$ 0.123 (1)} & 0.555 $\pm$ 0.098 (4) & 0.673 $\pm$ 0.081 (2) & 0.665 $\pm$ 0.147 (3) \\
colic & \xgbcell{0.768 $\pm$ 0.049} & \textbf{0.787 $\pm$ 0.055 (1)} & 0.772 $\pm$ 0.052 (3) & 0.766 $\pm$ 0.049 (4) & 0.775 $\pm$ 0.054 (2) \\
vote & \xgbcell{0.954 $\pm$ 0.019} & \textbf{0.955 $\pm$ 0.013 (1)} & 0.914 $\pm$ 0.038 (4) & 0.952 $\pm$ 0.018 (2) & 0.939 $\pm$ 0.025 (3) \\
breast-w & \xgbcell{0.973 $\pm$ 0.011} & \textbf{0.947 $\pm$ 0.011 (1)} & 0.927 $\pm$ 0.039 (4) & \textbf{0.947 $\pm$ 0.023 (1)} & 0.940 $\pm$ 0.015 (3) \\
blood-transfusion-service-center & \xgbcell{0.656 $\pm$ 0.032} & \textbf{0.666 $\pm$ 0.054 (1)} & 0.596 $\pm$ 0.033 (4) & 0.652 $\pm$ 0.016 (2) & 0.642 $\pm$ 0.041 (3) \\
diabetes & \xgbcell{0.789 $\pm$ 0.022} & 0.711 $\pm$ 0.038 (2) & 0.609 $\pm$ 0.105 (4) & 0.696 $\pm$ 0.042 (3) & \textbf{0.727 $\pm$ 0.028 (1)} \\
Mammographic-Mass-Data-Set & \xgbcell{0.799 $\pm$ 0.021} & 0.784 $\pm$ 0.031 (3) & 0.787 $\pm$ 0.013 (2) & 0.784 $\pm$ 0.008 (3) & \textbf{0.789 $\pm$ 0.015 (1)} \\
\hline
\multicolumn{6}{l}{\textbf{Medium datasets ($10^3 \leq n < 10^4$)}} \\
credit-g & \xgbcell{0.700 $\pm$ 0.041} & \textbf{0.703 $\pm$ 0.034 (1)} & 0.605 $\pm$ 0.041 (3) & 0.515 $\pm$ 0.023 (4) & 0.682 $\pm$ 0.017 (2) \\
qsar-biodeg & \xgbcell{0.868 $\pm$ 0.032} & \textbf{0.802 $\pm$ 0.034 (1)} & 0.759 $\pm$ 0.035 (4) & 0.777 $\pm$ 0.026 (3) & 0.785 $\pm$ 0.024 (2) \\
banknote-authentication & \xgbcell{0.999 $\pm$ 0.001} & \textbf{0.991 $\pm$ 0.005 (1)} & 0.960 $\pm$ 0.019 (3) & 0.970 $\pm$ 0.008 (2) & 0.958 $\pm$ 0.018 (4) \\
steel-plates-fault & \xgbcell{1.000 $\pm$ 0.000} & \textbf{0.947 $\pm$ 0.005 (1)} & 0.868 $\pm$ 0.077 (4) & 0.879 $\pm$ 0.015 (3) & \textbf{0.947 $\pm$ 0.005 (1)} \\
kr-vs-kp & \xgbcell{0.997 $\pm$ 0.002} & 0.927 $\pm$ 0.024 (3) & 0.882 $\pm$ 0.058 (4) & 0.940 $\pm$ 0.007 (2) & \textbf{0.942 $\pm$ 0.007 (1)} \\
spambase & \xgbcell{0.958 $\pm$ 0.008} & 0.865 $\pm$ 0.012 (3) & 0.823 $\pm$ 0.031 (4) & 0.891 $\pm$ 0.013 (2) & \textbf{0.892 $\pm$ 0.011 (1)} \\
phoneme & \xgbcell{0.873 $\pm$ 0.011} & \textbf{0.798 $\pm$ 0.009 (1)} & 0.650 $\pm$ 0.096 (4) & 0.757 $\pm$ 0.022 (3) & 0.794 $\pm$ 0.013 (2) \\
mushroom & \xgbcell{0.999 $\pm$ 0.001} & 0.992 $\pm$ 0.006 (3) & 0.989 $\pm$ 0.008 (4) & \textbf{0.998 $\pm$ 0.001 (1)} & 0.994 $\pm$ 0.004 (2) \\
\hline
\multicolumn{6}{l}{\textbf{Large datasets ($10^4 \leq n \leq 10^5$)}} \\
% eeg-eye-state & \xgbcell{0.941 $\pm$ 0.006} & \textbf{0.740 $\pm$ 0.004 (1)} & 0.675 $\pm$ 0.016 (4) & 0.684 $\pm$ 0.011 (3) & 0.690 $\pm$ 0.011 (2) \\
MagicTelescope & \xgbcell{0.865 $\pm$ 0.007} & \textbf{0.798 $\pm$ 0.002 (1)} & 0.753 $\pm$ 0.021 (4) & 0.773 $\pm$ 0.008 (2) & 0.754 $\pm$ 0.055 (3) \\
bank-marketing & \xgbcell{0.772 $\pm$ 0.017} & \textbf{0.807 $\pm$ 0.008 (1)} & 0.580 $\pm$ 0.080 (4) & 0.604 $\pm$ 0.058 (3) & 0.793 $\pm$ 0.005 (2) \\
electricity & \xgbcell{0.912 $\pm$ 0.006} & 0.774 $\pm$ 0.006 (2) & 0.737 $\pm$ 0.004 (3) & \textbf{0.793 $\pm$ 0.004 (1)} & 0.727 $\pm$ 0.048 (4) \\
adult & \xgbcell{0.780 $\pm$ 0.016} & \textbf{0.796 $\pm$ 0.008 (1)} & 0.687 $\pm$ 0.055 (3) & 0.554 $\pm$ 0.049 (4) & 0.785 $\pm$ 0.015 (2) \\
\hline
\multicolumn{6}{l}{\textbf{Massive datasets}} \\
creditcard & \xgbcell{0.925 $\pm$ 0.016} & \textbf{0.914 $\pm$ 0.010 (1)} & 0.774 $\pm$ 0.155 (4) & 0.865 $\pm$ 0.019 (3) & 0.909 $\pm$ 0.011 (2) \\
% Click\_prediction\_small & \xgbcell{0.669 $\pm$ 0.001} & 0.615 $\pm$ 0.010 (2) & 0.500 $\pm$ 0.000 (4) & 0.513 $\pm$ 0.003 (3) & \textbf{0.623 $\pm$ 0.003 (1)} \\
SEA(50000) & \xgbcell{0.836 $\pm$ 0.001} & \textbf{0.827 $\pm$ 0.004 (1)} & 0.769 $\pm$ 0.047 (4) & 0.810 $\pm$ 0.007 (3) & \textbf{0.827 $\pm$ 0.001 (1)} \\
\hline
Mean Bal. Acc.$\uparrow$ & \xgbcell{0.865} & \textbf{0.832} & 0.762 & 0.790 & 0.822 \\
Avg. Rank$\downarrow$ & \xgbcell{--} & \textbf{1.48} & 3.67 & 2.52 & 2.14 \\ 
Wins$\uparrow$ & \xgbcell{--} & \textbf{15} & 0 & 3 & 6 \\ 
\hline
\end{tabular}%
}
\end{table*}

\paragraph{Shared evaluation protocol}

To evaluate models in the shallow-tree regime, we constrain the maximum depth to $D\leq4$ for all single-tree methods. The baseline trees are binary by construction, whereas MBNDT tunes its maximum branching factor over $B\in\{3,4\}$. For the remaining model-specific hyperparameters, we follow the corresponding papers where possible; the full search spaces are reported in Tables~\ref{tab:appendix_hpo_space}--\ref{tab:appendix_hpo_xgb}.
We use nested stratified evaluation with fixed seeds and matching outer and
inner split indices across methods. 

\emph{Outer split:} for each dataset, five
80/20 train--test splits are generated, and each outer test set remains
untouched until final evaluation. 

\emph{Inner HPO:} within each outer-training
pool, Optuna evaluates configurations over five stratified inner folds under a
shared wall-clock budget of up to one hour per method and outer split. Each
configuration is scored by $\mu-0.5\sigma$, where $\mu$ and $\sigma$ are the
mean and standard deviation of inner-validation balanced accuracy. Within an
inner fold, differentiable methods further reserve 20\% of the inner-training
portion for early stopping, yielding a 64/16/20 parameter-training/early-stop/
inner-validation partition; non-neural methods train on the full 80\%
inner-training portion because they do not require early stopping.

\emph{Method-specific fitting:} all methods share the splits, depth constraint,
and wall-clock budget, but use their required preprocessing and optimization.
In particular, SPLIT uses quantile-binarized numeric features, one-hot encoded
categoricals, bounded lookahead, and per-solve time limits, with exact settings
reported in Table~\ref{tab:appendix_hpo_split}. 

\emph{Refit and test:} the best
configuration is refit on the outer-training pool (with an early-stopping
reserve only for differentiable methods) and evaluated once on the untouched
outer test set. Experiments ran on Ubuntu 24.04.2 LTS with an AMD Ryzen
Threadripper PRO 5975WX CPU and an NVIDIA RTX 4090 GPU.

\paragraph{MBNDT-specific training details}
For MBNDT, each Optuna trial samples only the architectural and optimizer
hyperparameters listed in Table~\ref{tab:appendix_hpo_space}. All remaining
training choices are fixed across datasets: Adam optimization with separate
learning-rate groups, balanced binary cross-entropy and deterministic batch-size
selection based on training-set size. Because differentiable tree training is sensitive to initialization, MBNDT uses
random restarts and checkpoint selection during the final refit. These choices
use only the early-stopping split drawn from the outer-training pool; the outer
test split is reserved for final reporting. Appendix~\ref{app:mbndt_training}
gives the exact MBNDT random restart policy and leaf-budget update.

\begin{table*}[!t]
\caption{Structural complexity and training time for depth-constrained
single-tree methods. Path length, realized leaves,
and time are averaged over five outer splits, with size-stratified means.}
\label{tab:appendix_complexity}
\centering
\scriptsize
\setlength{\tabcolsep}{6pt}
\renewcommand{\arraystretch}{0.90}
\resizebox{\textwidth}{!}{%
\begin{tabular}{l ccc@{\hspace{1.0em}}!{\color{black!35}\vrule width 0.3pt}@{\hspace{1.0em}}ccc@{\hspace{1.0em}}!{\color{black!35}\vrule width 0.3pt}@{\hspace{1.0em}}ccc@{\hspace{1.0em}}!{\color{black!35}\vrule width 0.3pt}@{\hspace{1.0em}}ccc}
\hline
Dataset & \multicolumn{3}{c}{MBNDT (ours)} & \multicolumn{3}{c}{GradTree} & \multicolumn{3}{c}{SPLIT} & \multicolumn{3}{c}{CART} \\
\cline{2-4}\cline{5-7}\cline{8-10}\cline{11-13}
 & Path & Leaves & Time (s) & Path & Leaves & Time (s) & Path & Leaves & Time (s) & Path & Leaves & Time (s) \\
\hline
\multicolumn{13}{l}{\textbf{Small datasets ($n < 10^3$)}} \\
hepatitis & 2.40 & 6.8 & 10.73 & 2.80 & 8.0 & 7.55 & 2.94 & 7.0 & 1.07 & 3.55 & 9.4 & $<0.01$ \\
colic & 1.66 & 4.0 & 19.81 & 3.00 & 8.8 & 12.75 & 1.92 & 4.6 & 0.70 & 3.36 & 9.2 & $<0.01$ \\
vote & 2.29 & 5.6 & 23.27 & 2.40 & 6.8 & 7.17 & 1.10 & 2.4 & 0.74 & 2.85 & 9.0 & $<0.01$ \\
breast-w & 2.62 & 18.2 & 20.33 & 3.40 & 11.2 & 7.67 & 2.07 & 5.2 & 0.94 & 3.33 & 10.6 & $<0.01$ \\
blood-transfusion-service-center & 2.26 & 9.0 & 15.45 & 3.60 & 12.8 & 16.83 & 2.46 & 6.4 & 2.26 & 2.48 & 7.4 & $<0.01$ \\
diabetes & 2.52 & 11.8 & 21.52 & 1.80 & 5.2 & 13.56 & 2.67 & 6.8 & 4.35 & 3.33 & 9.0 & $<0.01$ \\
Mammographic-Mass-Data-Set & 2.52 & 7.6 & 26.56 & 3.40 & 12.0 & 12.06 & 2.20 & 4.6 & 0.48 & 3.32 & 8.6 & $<0.01$ \\
\rowcolor{black!6}\textit{Small mean} & 2.32 & 9.0 & 19.67 & 2.91 & 9.3 & 11.08 & 2.19 & 5.3 & 1.51 & 3.18 & 9.0 & $<0.01$ \\
\hline
\multicolumn{13}{l}{\textbf{Medium datasets ($10^3 \leq n < 10^4$)}} \\
credit-g & 2.69 & 8.2 & 23.91 & 3.40 & 12.0 & 8.48 & 2.68 & 6.4 & 0.52 & 3.12 & 9.8 & $<0.01$ \\
qsar-biodeg & 3.36 & 18.6 & 21.86 & 3.80 & 14.4 & 27.61 & 2.63 & 6.8 & 3.87 & 3.57 & 10.6 & 0.01 \\
banknote-authentication & 2.99 & 37.4 & 27.63 & 4.00 & 16.0 & 9.65 & 2.93 & 7.4 & 11.20 & 3.50 & 11.6 & $<0.01$ \\
steel-plates-fault & 3.51 & 11.2 & 25.34 & 3.80 & 14.4 & 13.06 & 2.63 & 4.0 & 2.37 & 2.90 & 5.0 & $<0.01$ \\
kr-vs-kp & 3.20 & 9.0 & 22.25 & 3.80 & 14.4 & 14.64 & 2.69 & 6.0 & 1.16 & 3.07 & 6.6 & 0.01 \\
spambase & 3.60 & 17.6 & 28.61 & 3.60 & 13.6 & 21.75 & 3.00 & 8.0 & 26.92 & 3.95 & 13.2 & 0.02 \\
phoneme & 3.38 & 38.8 & 41.66 & 2.40 & 6.8 & 29.27 & 2.87 & 7.6 & 12.98 & 3.96 & 12.8 & 0.02 \\
mushroom & 3.28 & 8.6 & 43.54 & 4.00 & 16.0 & 23.94 & 2.95 & 7.0 & 5.54 & 3.74 & 10.0 & 0.01 \\
\rowcolor{black!6}\textit{Medium mean} & 3.25 & 18.7 & 29.35 & 3.60 & 13.5 & 18.55 & 2.80 & 6.7 & 8.07 & 3.48 & 10.0 & 0.01 \\
\hline
\multicolumn{13}{l}{\textbf{Large datasets ($10^4 \leq n \leq 10^5$)}} \\
% eeg-eye-state & 3.81 & 106.2 & 79.88 & 4.00 & 16.0 & 51.47 & 3.00 & 8.0 & 94.72 & 3.99 & 13.4 & 0.04 \\
MagicTelescope & 3.60 & 77.0 & 75.55 & 3.80 & 14.4 & 35.30 & 3.00 & 7.8 & 260.12 & 3.98 & 14.8 & 0.09 \\
bank-marketing & 3.30 & 19.0 & 126.50 & 3.60 & 12.8 & 115.27 & 2.99 & 6.4 & 5.83 & 3.34 & 10.6 & 0.08 \\
electricity & 3.51 & 49.0 & 169.21 & 3.00 & 10.0 & 180.12 & 3.00 & 8.0 & 180.13 & 3.49 & 11.2 & 0.04 \\
adult & 3.26 & 22.8 & 137.69 & 3.20 & 10.4 & 145.78 & 2.71 & 4.4 & 18.36 & 3.98 & 14.4 & 0.05 \\
\rowcolor{black!6}\textit{Large mean} & 3.42 & 42.0 & 127.24 & 3.40 & 11.9 & 119.12 & 2.92 & 6.7 & 116.11 & 3.70 & 12.8 & 0.07 \\
\hline
\multicolumn{13}{l}{\textbf{Massive datasets}} \\
creditcard & 2.75 & 16.6 & 633.69 & 3.60 & 12.8 & 315.87 & 1.42 & 4.6 & 251.50 & 2.98 & 6.0 & 6.24 \\
% Click\_prediction\_small & 3.09 & 19.0 & 729.36 & 2.00 & 4.8 & 358.78 & 1.61 & 4.8 & 1618.13 & 3.94 & 13.8 & 1.84 \\
SEA(50000) & 3.30 & 35.4 & 1616.55 & 2.80 & 9.2 & 1502.11 & 2.46 & 6.2 & 8974.47 & 3.94 & 15.8 & 1.51 \\
\rowcolor{black!6}\textit{Massive mean} & 3.03 & 26.0 & 1125.12 & 3.20 & 11.0 & 908.99 & 1.94 & 5.4 & 4612.99 & 3.46 & 10.9 & 3.88 \\
\hline
\rowcolor{black!6}\textit{Overall mean} & 2.95 & 20.6 & 149.13 & 3.30 & 11.5 & 120.02 & 2.54 & 6.1 & 465.02 & 3.42 & 10.3 & 0.39 \\
\hline
\end{tabular}%
}
\end{table*}

\subsection{Depth-constrained predictive performance}

Table~\ref{tab:per_dataset_balacc} reports mean and standard deviation of test
balanced accuracy over the five outer splits. Ranks and boldface are computed
over the single-tree methods only, with ties at three-decimal precision sharing
competition ranks. Among the depth-constrained single-tree learners,
MBNDT achieves the best average rank ($1.48$), the most wins ($15$ of $21$
datasets), and the highest mean balanced accuracy ($0.832$, versus $0.822$ for
CART, $0.790$ for SPLIT, and $0.762$ for GradTree). A Friedman test rejects
equal performance across the four single-tree methods ($p=2.37\times10^{-6}$),
and pairwise two-sided Wilcoxon signed-rank tests with Holm correction confirm
that MBNDT improves over each baseline individually (Table~\ref{tab:paired_tests}).
The margin is largest over GradTree and SPLIT and smaller---though still
significant---over greedy CART, consistent with CART being the strongest of the
depth-constrained baselines.

\begin{table}[h]
\caption{Paired Wilcoxon tests for test balanced accuracy between MBNDT and
single-tree baselines with maximum depth $D\leq4$.}
\label{tab:paired_tests}
\centering
\scriptsize
\setlength{\tabcolsep}{6pt}
\resizebox{\columnwidth}{!}{%
\begin{tabular}{lccccc}
\hline
Comparison & $N$ & W/T/L & Wilcoxon $p$ & Holm $p$ \\
\hline
MBNDT vs. GradTree & 21 & 20/0/1  & $2.86\times 10^{-6}$ & $8.58\times 10^{-6}$ \\
MBNDT vs. SPLIT & 21 & 15/2/4  & 0.00628 & 0.01256 \\
MBNDT vs. CART & 21 & 14/2/5  & 0.02633 & 0.02633 \\
\hline
\end{tabular}%
}
\end{table}

\subsection{Structural complexity and trade-offs}

Table~\ref{tab:appendix_complexity} reports mean test decision-path length,
realized leaves, and final refit time after model selection, averaged over the
five outer splits and broken out by dataset size. We treat path length as a
proxy for local decision cost and realized leaves as a proxy for global model
size; neither metric alone establishes human interpretability. These results
make the intended trade-off explicit. MBNDT keeps per-instance decision paths short
(mean $2.95$, shorter than GradTree and CART, though longer than SPLIT) while
realizing more leaves ($20.6$) than the binary or solver-based baselines. SPLIT
produces the smallest trees but the lowest accuracy (Table~\ref{tab:per_dataset_balacc}).
MBNDT's refit time is higher than CART and GradTree but lower than SPLIT on
average.

\subsection{Ablation studies of MBNDT}

The ablations are diagnostic rather than alternate main benchmarks: the main
tables use HPO-selected MBNDT with the leaf-budget penalty enabled, whereas
Tables~\ref{tab:branch_factor_ablation}--\ref{tab:maskpp} fix or remove that
budget to isolate branching, masks, and post-hoc pruning.

\begin{table}[h]
\caption{Branching-factor ablation for MBNDT with fixed depth $D=4$ and
leaf budget $K=16$. Best counts the number of datasets on which each variant
achieves the highest balanced accuracy.}
\label{tab:branch_factor_ablation}
\centering
\scriptsize
\setlength{\tabcolsep}{3pt}
\begin{tabular}{lccccc}
\hline
Variant & $B$ & Bal. Acc. & Path & Leaves & Best \\
\hline
MBNDT-B2 & 2 & 0.820 & 3.22 & \textbf{9.5} & 4 \\
MBNDT-B3 & 3 & \textbf{0.833} & 3.12 & 16.0 & \textbf{10} \\
MBNDT-B4 & 4 & 0.832 & \textbf{3.09} & 23.3 & 7 \\
\hline
\end{tabular}
\end{table}

\paragraph{Effect of branching factor.}

To isolate the contribution of multi-way splitting, we run a fixed-hyperparameter branch-factor ablation with $B \in \{2,3,4\}$. For each outer split, we reuse the non-architecture hyperparameters from the original HPO run, including learning rates and routing temperature, and hold the depth and soft leaf budget fixed at $D=4$ and $K=16$. This is therefore not a full HPO comparison across branching factors; it tests how local branching capacity affects performance under a common training and budget setting.

Table~\ref{tab:branch_factor_ablation} reports mean balanced accuracy, mean test path length, realized leaves, and the number of datasets on which each variant is best among the three branching factors. Moving from binary to ternary routing raises balanced accuracy from $0.820$ to $0.833$ and slightly shortens mean path length from $3.22$ to $3.12$; in the direct pairwise comparison, B3 outperforms B2 on 16 of 21 datasets. This suggests that the improvement is not driven by longer decision paths, but by greater local partitioning capacity at fixed depth. The cost is increased realized leaves ($9.5$ to $16.0$). Increasing to $B=4$ gives similar accuracy ($0.832$) and the shortest paths ($3.09$), but realizes more leaves ($23.3$), indicating diminishing returns beyond ternary splits under this budget.

\begin{table}[h]
\caption{Effect of branch masks and post-hoc train-path pruning (PP) in a
controlled setting without a leaf-budget penalty.}
\label{tab:maskpp}
\centering
\scriptsize
\setlength{\tabcolsep}{4pt}
\begin{tabular}{ccccc}
\hline
Masks & PP & Bal. Acc. $\uparrow$ & Path $\downarrow$ & Leaves $\downarrow$ \\
\hline
\ding{55} & \ding{55} & 0.823 & 3.32 & 60.7 \\
\ding{51} & \ding{55} & 0.827 & 3.33 & 56.4 \\
\ding{55} & \ding{51} & 0.823 & 3.06 & 22.9 \\
\ding{51} & \ding{51} & \textbf{0.827} & \textbf{3.04} & \textbf{20.7} \\
\hline
\end{tabular}
\end{table}

\paragraph{Effect of branch masks.}
Table~\ref{tab:maskpp} separates branch masks from post-hoc train-path
pruning (PP) in a controlled no-budget setting with fixed hyperparameters;
the leaf-budget penalty is disabled because its surrogate depends on the
masked routing probabilities (Eq.~\ref{eq:lsoft}) and would confound the
mask-versus-PP attribution. PP is the main
compression mechanism: without masks, it reduces realized leaves from $60.7$
to $22.9$ on average ($p<10^{-4}$) and shortens mean paths from $3.32$ to
$3.06$, while balanced accuracy is statistically unchanged ($0.823$,
$p=0.13$). Branch masks play a different role. Without PP, masks reduce
leaves only modestly ($60.7$ to $56.4$) but raise balanced accuracy from
$0.823$ to $0.827$ (paired Wilcoxon $p=0.010$). After applying the same PP
step, the masked model remains more accurate ($0.827$ vs.\ $0.823$;
$p=0.025$, 15/21 datasets) and slightly smaller ($20.7$ vs.\ $22.9$ leaves).

Thus, masks are not simply a substitute for post-hoc pruning: PP supplies
most of the size reduction, whereas masks shape the routing during training
and yield a small but consistent accuracy gain that survives pruning.
Overall, their combination gives the most favorable accuracy--size
trade-off, reducing realized leaves by about two-thirds relative to the
unmasked, unpruned tree.

\paragraph{Effect of the leaf-budget constraint.}

Table~\ref{tab:leaf_budget_ablation} reports a fixed-budget sweep against
single-tree reference points; MBNDT values are averaged over datasets after
post-hoc pruning. The leaf budget behaves as an effective accuracy--size
control knob: increasing it from $K=4$ to $K=64$ raises mean balanced
accuracy from $0.827$ to $0.836$ while lengthening mean paths from $2.67$ to
$3.01$ and realized leaves from $13.4$ to $20.3$.

\begin{table}[h]
\caption{Leaf-budget ablation for MBNDT with single-tree reference points.
MBNDT rows vary the target leaf budget $K$; the HPO-selected row reports the
mean selected $K$ across dataset splits. Reference methods are shown under the
same depth constraint.}
\label{tab:leaf_budget_ablation}
\centering
\scriptsize
\setlength{\tabcolsep}{3pt}
\resizebox{\columnwidth}{!}{%
\begin{tabular}{lcccc}
\hline
Method / setting & Target $K$ & Bal. Acc. & Path & Leaves \\
\hline
SPLIT & -- & 0.790 & \textbf{2.54} & \textbf{6.1} \\
CART & -- & 0.822 & 3.42 & 10.3 \\
GradTree & -- & 0.762 & 3.30 & 11.5 \\
\hline
MBNDT, tight budget & 4 & 0.827 & 2.67 & 13.4 \\
MBNDT, moderate budget & 16 & 0.835 & 2.94 & 18.3 \\
MBNDT, large budget & 64 & \textbf{0.836} & 3.01 & 20.3 \\
MBNDT, no budget & -- & 0.827 & 3.04 & 20.7 \\
MBNDT, HPO-selected & 31.0 & 0.832 & 2.95 & 20.6 \\
\hline
\end{tabular}
}
\end{table}

The effect saturates and
is not strictly monotonic across datasets---$K=16$ already attains $0.835$
accuracy at path $2.94$ and $18.3$ leaves, close to $K=64$. The penalty is
moreover regularizing rather than merely shrinking: on the dataset-averaged
means, $K=16$ dominates the no-budget variant on all three axes (higher
accuracy, shorter paths, and fewer leaves than $0.827$/$3.04$/$20.7$).
Relative to the single-tree baselines, MBNDT occupies a higher-accuracy,
larger-tree regime: SPLIT, CART, and GradTree remain more compact ($6.1$,
$10.3$, and $11.5$ leaves) but less accurate ($0.790$, $0.822$, and $0.762$),
and even the tightest budget ($K=4$, $13.4$ leaves) does not bring MBNDT below
them in size. The
budget should therefore be read as a knob for trading realized size against
accuracy within MBNDT, not as a mechanism that makes it globally smaller
than binary or solver-based trees. (The no-budget row matches the setting of
the mask/PP ablation in Table~\ref{tab:maskpp}; the HPO-selected row
corresponds to the main benchmark protocol.)

\section{Conclusion}

We studied shallow-depth tree induction, where the complexity budget is a maximum decision depth, and proposed MBNDT, a single axis-aligned tree that learns differentiable multi-way splits end-to-end. By
increasing local partitioning capacity instead of tree depth, MBNDT achieves
the highest mean balanced accuracy among depth-constrained single-tree learners
across 21 OpenML binary-classification benchmarks. A branching-factor ablation
shows that multi-way splits outperform a binary variant under the same depth and
budget setting, supporting the role of multi-way branching. The improvement has
an explicit cost---MBNDT realizes more leaves than the other single-tree
baselines---so it is best suited to settings that prioritize accuracy under
bounded per-instance decision depth over minimal global tree size.

Several directions remain for future work. First, MBNDT is evaluated here for
binary classification; extending the differentiable multi-way formulation to
multiclass classification and regression is a natural next step. In addition,
future work should complement path length and realized leaf count with
distinct-feature and threshold/rule-complexity measures and human-centered
studies.

\section*{Acknowledgment}

This work was supported by the Institute of Information \& Communications Technology Planning \& Evaluation (IITP) under the Artificial Intelligence Convergence Innovation Human Resources Development (IITP-2026-RS-2023-00254177), the Institute of Information \& Communications Technology Planning \& Evaluation (IITP) under the Leading Generative AI Human Resources Development (IITP-2026-RS-2024-00360227), and the National Research Foundation of Korea (NRF) grant funded by the Korean government (MSIT) (RS-2026-25555206).

\bibliographystyle{IEEEtran}
\bibliography{references}

\clearpage
\appendix

\subsection{MBNDT training details}
\label{app:mbndt_training}

\paragraph{Restart policy}
Final refits use five random restarts. Stage~1 is run from each initialization
and the restart checkpoint is selected using validation loss; stage~2 resumes
from the selected checkpoint and early-stops on validation balanced accuracy.
During the final refit, the corresponding limits are $40/8$ and $500/25$
epochs/patience. 

\paragraph{Leaf-budget update}
For the selected leaf budget $K$, the experiments use the log-scale violation
$v=[\log L_{\mathrm{soft}}-\log K]_+$. The minibatch objective adds
$\mu v+(\rho/2)v^2$ with fixed $\rho=0.03$ and initial multiplier $\mu_0=0$.
After each epoch, the mean training-set violation is accumulated with an
exponential moving average using coefficient $0.5$, and the multiplier is
updated by projected dual ascent,
$\mu\leftarrow\max\{0,\mu+0.003\,\widehat{v}\}$, where $\widehat{v}$ denotes
the smoothed violation.

\subsection{Dataset characteristics and preprocessing detail}

\begin{table}[h]
\caption{Dataset characteristics for the 21 OpenML binary-classification benchmarks.}
\label{tab:appendix_datasets}
\centering
\tiny
\setlength{\tabcolsep}{3pt}
\renewcommand{\arraystretch}{0.70}
\resizebox{\columnwidth}{!}{%
\begin{tabular}{l r r r}
\hline
Dataset & OpenML ID & Instances & Attributes \\
\hline
hepatitis & 55 & 155 & 20 \\
colic & 27 & 368 & 23 \\
vote & 56 & 435 & 17 \\
breast-w & 15 & 699 & 10 \\
blood-transfusion-service-center & 1464 & 748 & 5 \\
diabetes & 37 & 768 & 9 \\
Mammographic-Mass-Data-Set & 45557 & 961 & 5 \\
credit-g & 31 & 1{,}000 & 21 \\
qsar-biodeg & 1494 & 1{,}055 & 42 \\
banknote-authentication & 1462 & 1{,}372 & 5 \\
steel-plates-fault & 1504 & 1{,}941 & 34 \\
kr-vs-kp & 3 & 3{,}196 & 37 \\
spambase & 44 & 4{,}601 & 58 \\
phoneme & 1489 & 5{,}404 & 6 \\
mushroom & 24 & 8{,}124 & 23 \\
% eeg-eye-state & 1471 & 14{,}980 & 15 \\
MagicTelescope & 1120 & 19{,}020 & 12 \\
bank-marketing & 1461 & 45{,}211 & 17 \\
electricity & 151 & 45{,}312 & 9 \\
adult & 1590 & 48{,}842 & 15 \\
creditcard & 1597 & 284{,}807 & 31 \\
% Click\_prediction\_small & 1219 & 399{,}482 & 12 \\
SEA(50000) & 162 & 1{,}000{,}000 & 4 \\
\hline
\end{tabular}%
}
\end{table}

Table~\ref{tab:appendix_datasets} summarizes the publicly available OpenML datasets used in the main experiments. All preprocessing is leakage-safe, fitted on training data only. Numeric features are median-imputed. Categorical features with more than ten unique
values use leave-one-out (LOO) target encoding, and lower-cardinality features
use one-hot encoding; LOO transforms use labels only when fitting on training
data, while validation and test transforms are label-free, with unseen or
missing categories mapped to a default ``unknown'' value. For the
differentiable methods (MBNDT, GradTree), heavy-tailed numeric features
optionally receive a $\log(1+x)$ transform followed by rank--Gaussian
normalization to $\mathcal{N}(0,1)$, which stabilizes differentiable
threshold learning; these transforms are method-appropriate and do not alter
the split structure available to the other tree baselines. For SPLIT, numeric
features are quantile-discretized and categoricals one-hot encoded, as
required by its binarized solver.

\subsection{Model-wise hyperparameter search space}

\begin{table}[h]
\caption{MBNDT hyperparameter search space.}
\label{tab:appendix_hpo_space}
\centering
\scriptsize
\setlength{\tabcolsep}{2pt}
\renewcommand{\arraystretch}{0.86}
\begin{tabular}{@{}p{0.20\columnwidth}p{0.66\columnwidth}p{0.06\columnwidth}@{}}
\hline
Hyperparameter & Values & Scale \\
\hline
$B$ & $\{3,4\}$ & cat. \\
$D$ & $\{1,2,3,4\}$ & cat. \\
$K\mid B=3$ & \begin{tabular}[t]{@{}l@{\hspace{0.25em}}l@{}}$D_1:\{2,3\}$ & $D_2:\{3,6,9\}$ \\$D_3:\{6,12,18,24\}$ & $D_4:\{12,24,36,48\}$\end{tabular} & cat. \\
$K\mid B=4$ & \begin{tabular}[t]{@{}l@{\hspace{0.25em}}l@{}}$D_1:\{2,3,4\}$ & $D_2:\{4,8,12,16\}$ \\$D_3:\{8,12,16,24,32\}$ & $D_4:\{16,24,32,48,64\}$\end{tabular} & cat. \\
$\tau_{\mathrm{cdf}}$ & $[0.08,1.2]$ & log \\
$\eta_*$ & $[10^{-3},10^{-1}]$ for feature, threshold, leaf, mask & log \\
Loss & balanced BCE & fixed \\
\hline
\end{tabular}
\end{table}

\begin{table}[h]
\caption{GradTree hyperparameter search space.}
\label{tab:appendix_hpo_gradtree}
\centering
\scriptsize
\setlength{\tabcolsep}{2pt}
\renewcommand{\arraystretch}{0.86}
\begin{tabular}{@{}p{0.30\columnwidth}p{0.50\columnwidth}p{0.12\columnwidth}@{}}
\hline
Hyperparameter & Values & Scale \\
\hline
$D$ & $\{1,2,3,4\}$ & cat. \\
$\eta_{\mathrm{index}}$ & $[10^{-3},10^{-1}]$ & log \\
$\eta_{\mathrm{value}}$ & $[10^{-3},10^{-1}]$ & log \\
$\eta_{\mathrm{leaf}}$ & $[10^{-3},10^{-1}]$ & log \\
Focal loss & $\{0,1\}$ & cat. \\
polyLoss & $\{0,1\}$ & cat. \\
polyLossEpsilon & $\{0,1,2,3,4,5\}$ & int \\
Epochs / patience / batch & $250 / 25 / 128$ & fixed \\
SWA / class balance & $1 / 1$ & fixed \\
Initialization & normal & fixed \\
\hline
\end{tabular}
\end{table}

\begin{table}[h]
\caption{SPLIT hyperparameter search space.}
\label{tab:appendix_hpo_split}
\centering
\scriptsize
\setlength{\tabcolsep}{2pt}
\renewcommand{\arraystretch}{0.86}
\begin{tabular}{@{}p{0.30\columnwidth}p{0.50\columnwidth}p{0.12\columnwidth}@{}}
\hline
Hyperparameter & Values & Scale \\
\hline
Regularization & $[10^{-6},10^{-1}]$ & log \\
full\_depth\_budget & $4$ & fixed \\
Lookahead, small folds & $3$ & fixed \\
Lookahead, other folds & $2$ & fixed \\
Binarizer estimators & $50$ & fixed \\
Binarizer depth & $1$ & fixed \\
HPO / final fit limit & $300\,\mathrm{s} / 1200\,\mathrm{s}$ & fixed \\
\hline
\end{tabular}
\end{table}

\begin{table}[h]
\caption{CART hyperparameter search space.}
\label{tab:appendix_hpo_cart}
\centering
\scriptsize
\setlength{\tabcolsep}{2pt}
\renewcommand{\arraystretch}{0.86}
\begin{tabular}{@{}p{0.30\columnwidth}p{0.50\columnwidth}p{0.12\columnwidth}@{}}
\hline
Hyperparameter & Values & Scale \\
\hline
criterion & $\{\mathrm{gini},\mathrm{entropy},\mathrm{log\_loss}\}$ & cat. \\
splitter & $\{\mathrm{best},\mathrm{random}\}$ & cat. \\
max\_depth & $\{1,2,3,4\}$ & cat. \\
min\_samples\_split & $[10^{-3},0.20]$ & log \\
min\_samples\_leaf & $[10^{-3},0.20]$ & log \\
max\_features & $\{\mathrm{None},\mathrm{sqrt},\mathrm{log2}\}$ & cat. \\
class\_weight & $\{\mathrm{None},\mathrm{balanced}\}$ & cat. \\
ccp\_alpha & $[10^{-8},10^{-2}]$ & log \\
\hline
\end{tabular}
\end{table}

\begin{table}[h]
\caption{XGBoost hyperparameter search space.}
\label{tab:appendix_hpo_xgb}
\centering
\scriptsize
\setlength{\tabcolsep}{2pt}
\renewcommand{\arraystretch}{0.86}
\begin{tabular}{@{}p{0.30\columnwidth}p{0.50\columnwidth}p{0.12\columnwidth}@{}}
\hline
Hyperparameter & Values & Scale \\
\hline
n\_estimators & $[50,500]$ & int \\
max\_depth & $\{1,2,3,4\}$ & cat. \\
learning\_rate & $[10^{-2},3\times10^{-1}]$ & log \\
min\_child\_weight & $[10^{-2},20]$ & log \\
gamma & $[10^{-8},5]$ & log \\
reg\_alpha & $[10^{-8},10]$ & log \\
reg\_lambda & $[10^{-3},100]$ & log \\
subsample & $[0.5,1.0]$ & linear \\
column sampling & $[0.5,1.0]$ & linear \\
scale\_pos\_weight mode & $\{\mathrm{none},\mathrm{balanced}\}$ & cat. \\
\hline
\end{tabular}
\end{table}

\end{document}